%% file: main.tex
\documentclass{article}
\usepackage[preprint]{arxiv}
\usepackage[utf8]{inputenc} 
\usepackage[T1]{fontenc}    
\usepackage[hidelinks]{hyperref}       
\usepackage{url}            
\usepackage{booktabs}       
\usepackage{amsfonts}       
\usepackage{amssymb}        
\usepackage{mathtools}      
\usepackage{nicefrac}       
\usepackage{microtype}      
\usepackage{xcolor}         
\usepackage{cleveref}       
\usepackage{lipsum}         
\usepackage{graphicx}       
\usepackage{doi}
\usepackage{authblk}
\usepackage[printonlyused, withpage]{acronym}        
\usepackage[english]{babel}
\usepackage{siunitx} 
\usepackage[frozencache,cachedir=.]{minted} 

\input{sections/preamble}

\newacro{dl}[DL]{deep learning}
\newacro{fm}[FM]{foundation model}
\newacro{sfm}[SFM]{seismic foundation model}
\newacro{vit}[ViT]{Vision Transformer}
\newacro{swin}[Swin]{Swin Transformer}
\newacro{davit}[DaViT]{Dual Attention Vision Transformer}
\newacro{psnr}[PSNR]{peak signal-to-noise ratio}
\newacro{ssim}[SSIM]{structural similarity index measure}
\newacro{mse}[MSE]{mean squared error}
\newacro{bert}[BERT]{Bidirectional Encoder Representations from Transformers}
\newacro{mae}[MAE]{masked autoencoders}
\newacro{gsfm}[GSFM]{generative seismic foundation model}
\newacro{mim}[MIM]{Masked Image Modeling}
\newacro{cnn}[CNN]{convolutional neural network}
\newacro{nn}[NN]{neural network}
\newacro{cv}[CV]{computer vision}

\title{Foundation Models For Seismic Data Processing: An Extensive Review}

\newbox{\orcid}\sbox{\orcid}{\includegraphics[scale=0.06]{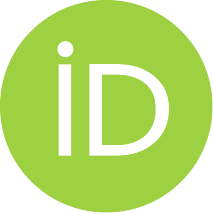}}
\author[1,2]{%
	\href{https://orcid.org/0009-0009-1060-6093}{\usebox{\orcid}\hspace{1mm}Fabian Fuchs\thanks{\texttt{fabian.fuchs@itwm.fraunhofer.de}}}
}
\author[1]{%
	\href{https://orcid.org/0009-0006-3945-1863}{\usebox{\orcid}\hspace{1mm}Mario Ruben Fernandez}%
}
\author[1]{%
	Norman Ettrich
}
\author[2,3]{%
	\href{https://orcid.org/0000-0002-1327-1243}{\usebox{\orcid}\hspace{1mm}Janis Keuper}%
}
\affil[1]{Fraunhofer-Institut für Techno- und Wirtschaftsmathematik}
\affil[2]{DWS, University of Mannheim}
\affil[3]{IMLA, Offenburg University}

\begin{document}

\maketitle

\input{sections/abstract}

\input{sections/introduction}
\input{sections/methodology}
\input{sections/results}
\input{sections/discussion}
\input{sections/conclusion}

\section{Acknowledgments}

The authors would like to acknowledge the members of the Fraunhofer ITWM DLSeis Consortium (http://dlseis.org) for their financial support. We are grateful to Equinor and Volve Licence partners for releasing The North Sea Volve Dataset under an Equinor Open Data Licence. We are also grateful to Hemang Shah and BP Exploration Operation Company Limited ("BP") for providing the 2007 BP Anisotropic Velocity Benchmark dataset.

\section{Data and Materials availability}

The training data for the interpolation and denoise task is the 2007 BP Anisotropic Velocity Benchmark dataset \cite[]{shah2007BPAnisotropic2007}. It is an open seismic dataset created by Hemang Shah and provided  by BP Exploration Operation Company Limited ("BP"). It can be obtained by visiting \url{https://wiki.seg.org/wiki/2007_BP_Anisotropic_Velocity_Benchmark}.
Additionally, the North Sea Volve dataset  was used for the qualitative demultiple results. It is also an open seismic dataset and can be obtained by visiting \url{https://www.equinor.com/energy/volvedata-sharing}.

The source code is available at: \url{https://codeberg.org/fuchsfa/foundation-models-seismic-processing}. 
In addition to the source code, the experimental metrics and settings are also saved in the Git repository and accessible via DVC. Together with the open source 2007 BP Anisotropic Velocity Benchmark dataset this enables the full reproduction of the interpolation and denoising models used in this study.

\bibliographystyle{unsrtnat}
\bibliography{ComputerVision}

\end{document}

%% file: sections/preamble.tex
\definecolor{CAFORMER}{HTML}{000000}
\definecolor{CONVNEXT}{HTML}{006400}
\definecolor{DAVIT}{HTML}{00008b}
\definecolor{DEIT3}{HTML}{b03060}
\definecolor{DINO}{HTML}{ff4500}
\definecolor{EFFICIENTNET}{HTML}{ffff00}
\definecolor{EVA02}{HTML}{00ff00}
\definecolor{HIERA}{HTML}{00ffff}
\definecolor{MAMBAOUT}{HTML}{ff00ff}
\definecolor{SWIN}{HTML}{6495ed}
\definecolor{VITAMIN}{HTML}{deb887}
\newcommand{\centeredcolorbox}[1]{ \textcolor{#1}{\rule{0.3cm}{0.3cm}}}
\newcommand{\trainingshape}[1]{\includegraphics[width=0.3cm]{images/#1}}
\newcommand{\convnext}{ConvNeXt}
\newcommand{\vitamin}{ViTamin}
\newcommand{\efficientnet}{EfficientNet}
\newcommand{\hiera}{Hiera}
\newcommand{\caformer}{CAFormer}
\newcommand{\imgnet}[1]{ImageNet-#1k}
\newcommand{\unet}{UNet}
\newcommand{\unets}{UNets}

%% file: sections/abstract.tex
\begin{abstract}
  Seismic processing plays a crucial role in transforming raw data into high-quality subsurface images, pivotal for various geoscience applications. Despite its importance, traditional seismic processing techniques face challenges such as noisy and damaged data and the reliance on manual, time-consuming workflows. The emergence of deep learning approaches has introduced effective and user-friendly alternatives, yet many of these deep learning approaches rely on synthetic datasets and specialized neural networks. 
  Recently, foundation models have gained traction in the seismic domain, due to their success in the natural image domain. 
  Therefore, we investigate the application of natural image foundation models on the three seismic processing tasks: demultiple, interpolation, and denoising. 
  We evaluate the impact of different model characteristics, such as pre-training technique and neural network architecture, on performance and efficiency. 
  Rather than proposing a single seismic foundation model, we critically examine various natural image foundation models and suggest some promising candidates for future exploration.

\end{abstract}

%% file: sections/introduction.tex
\section{Introduction}
\label{sec:introduction}

Seismic processing is essential for converting raw data into high-quality subsurface images. Precise subsurface imaging is vital for various geoscience applications, such as sedimentary and tectonic interpretation, hydrocarbon exploration, reservoir analysis, and geothermal characterization \cite[]{yilmazSeismicDataAnalysis2001a}.  
The challenges associated with this process include among others environmental noise, damaged geophones, and weak low-frequency signals, which degrade data accuracy. Therefore, this transformation involves complex processing steps, relying on time-consuming workflows that approximate the physics behind wave propagation, to enhance the quality of the data. 
Additionally, these techniques frequently require manual processes like picking velocities and mute functions, as well as parameter tuning, generally involving iterative adjustments based on the dataset. 
Therefore, seismic processing requires significant human expertise to extract geological and physical information while mitigating undesired artifacts inherent in the data or introduced by the processing.

In recent years, motivated by the success of \ac{dl} approaches in \ac{cv}, \ac{dl} based approaches have emerged as effective, parameter-free, and user-friendly alternatives to traditional seismic processing techniques \cite[]{mousaviDeeplearningSeismology2022}. Recent studies have successfully applied \ac{dl} based approaches to seismic data interpolation, denoising, deblending, ground-roll attenuation, velocity picking, and dispersion curve estimation (\cite{chaiDeepLearningIrregularly2020}, \cite{fernandezBENCHMARKSTUDYDEEP2022}, \cite{fernandezComparisonDeepLearning2022a}, \cite{mandelliInterpolationDenoisingSeismic2019}, \cite{birniePotentialSelfsupervisedNetworks2021}, \cite{qiuDeepLearningPrior2022}, \cite{zhangUnsupervisedSeismicRandom2019}, \cite{wangSeismicDeblendingSelfsupervised2021}, \cite{guoGroundRollAttenuation2020}, \cite{kaurSeismicGroundrollNoise2020}, \cite{wangAutomaticVelocityPicking2021}, \cite{chamorroDeepLearningbasedExtraction2024}). However, the majority of these \ac{dl} based solutions are specialized neural networks trained for a single specific task. 

In contrast, \acp{fm} are large neural networks that have been pre-trained on vast amounts of data in order to be used in a wide range of downstream tasks. This approach has recently led to numerous breakthroughs across diverse domains, including: ChatGPT \cite[]{brownLanguageModelsAre2020} in the natural language processing domain,  SAM \cite[]{kirillovSegmentAnything2023b} in image segmentation, Stable Diffusion \cite[]{rombachHighResolutionImageSynthesis2022} in image generation,  SORA \cite[]{liuSoraReviewBackground2024} in video generation and image encoders, such as \ac{vit} \cite[]{dosovitskiyImageWorth16x162020}, \convnext{}   \cite[]{liuConvNet2020s2022} and \ac{swin} \cite[]{liuSwinTransformerHierarchical2021}.

\begin{figure}
    \centering
    \includegraphics[width=\textwidth]{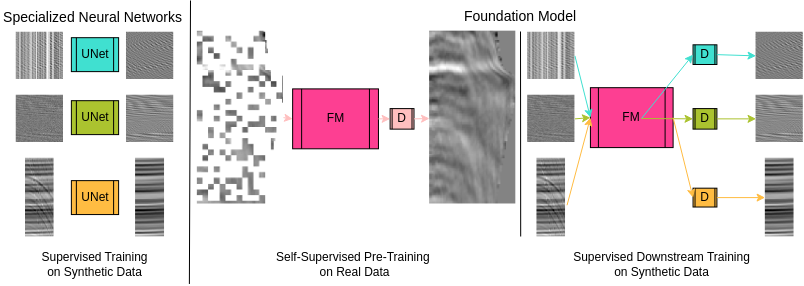}
    \caption{Comparison between specialized neural networks and \acp{fm}. Specialized neural networks, like \unets, are trained end to end for each downstream task. In comparison, an \ac{fm} is pre-trained once and then fine-tuned for each downstream task. The pre-training is done though self-supervised learning on real field data and the fine-tuning through supervised learning on synthetic data. The parts of the model denoted as D are task dependent decoders.}
    \label{fig:fm}
\end{figure}

The success of \ac{dl} based approaches in seismic and the achievements of \acp{fm} in \ac{cv} have prompted a surge of research interest in \acp{sfm}. Currently, most \ac{dl} approaches for seismic applications rely on synthetic data, due to the difficulties in labeling field seismic data. This presents a challenge, as synthetic datasets often fail to capture the true diversity and characteristics of field seismic data. A key benefit of \acp{sfm} is their ability to be pre-trained on and learn from seismic field data before being fine-tuned on synthetic data for the specific downstream tasks, as shown in Figure \ref{fig:fm}. The ability to learn directly from seismic field data, in combination with larger neural networks inherent to \ac{fm}, promises improved performance and even more importantly better generalization. 

\subsection{FMs in geophysics}
Although the term \ac{fm} is not explicitly used, StorSeismic from \cite{harsukoStorSeismicNewParadigm2022} is one of the first reported \ac{sfm}. StorSeismic utilizes a \ac{bert} model, as outlined in  \cite{devlinBERTPretrainingDeep2019}. It is first pre-trained on a mixture of real and synthetic data seismic data and subsequently fine-tuned for different downstream tasks.

In the field of seismic processing, StorSeismic employs a distinctive approach by treating each trace as a word, and a gather as a sentence. This strategy is supported by \ac{bert}, a model specifically engineered for natural language processing. In contrast, a more prevalent strategy within the seismic processing domain involves treating the gather as an image and leveraging models from the natural image domain. 
This approach is used by \cite{shengSeismicFoundationModel2024}, \cite{phamSeisBERTPretrainedSeismic2025} and \cite{sansalScalingSeismicFoundation2025}, who all utilize a \ac{vit} \cite[]{dosovitskiyImageWorth16x162020} as an image encoder. All three adopt the \ac{mae} framework \cite[]{heMaskedAutoencodersAre2021}, which trains an encoder-decoder model via self-supervision. This process entails masking part of the input gather and training the model to recreate the whole gather. Following this pre-training phase, the encoder component of the model is employed as a feature extractor for a range of downstream tasks.
A notable distinction between the approach presented by \cite{sansalScalingSeismicFoundation2025} and the other two image-based approaches is, that the former uses three-dimensional seismic data, which is a more complex problem. 

Another image-based approach is the \ac{gsfm} by \cite
{chengGenerativeFoundationModel2025}, which utilizes a diffusion model. In contrast, to the \acp{sfm} previously mentioned, their \ac{gsfm} is pre-trained and fine-tuned for all their downstream tasks simultaneously. This enables them to share the whole network, not just the image encoder, between downstream tasks.

\subsection{Motivation}

The majority of the image-based approaches previously referenced, except \cite{chengGenerativeFoundationModel2025}, use a \ac{vit} as their image encoder. While \acp{vit} have been extensively utilized in the domain of natural images, there is a need to assess their suitability for the domain of seismic processing. This is particularly salient in light of recent studies by \cite{goldblumBattleBackbonesLargeScale2023}, which empirically demonstrate that \acp{vit} may not be the optimal architecture even for all natural image applications. Therefore, there is a need for comparing and benchmarking other \acp{fm} for seismic processing.

In this paper, we conduct a comprehensive investigation into various \acp{fm} and develop a framework for further testing. We assess the performance of the different \acp{fm} on three seismic processing tasks: demultiple, interpolation, and denoising. For the latter two tasks, we utilize an open-source dataset, which, when combined with the source code that will be made available, allows for the reproduction of our results and serves as a foundation for further research. We quantitatively evaluate a diverse range of \acp{fm} using synthetic data, aiming to clarify how various model characteristics, such as pre-training techniques and architectural design choices, impact their performance. Additionally, we select a subset of models for qualitative assessment, where we present demultiple results of real field data. Finally, we aim to investigate the impact of natural image pre-training, because unlike curated seismic data sets, extensive natural image datasets are widely available.

%% file: sections/methodology.tex
\section{Methodology}

\begin{figure}
    \centering
    \includegraphics[width=\textwidth]{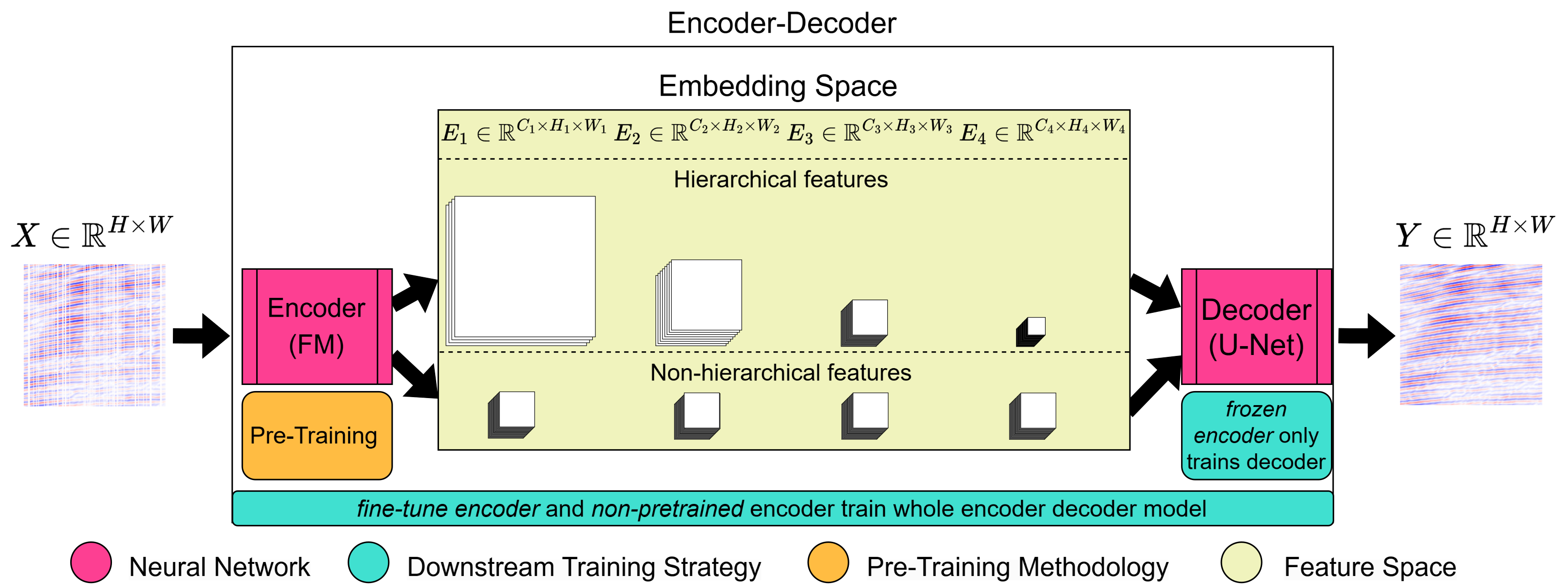}
    \caption{Depiction of the architecture of the encoder-decoder model with an arbitrary \ac{fm}, that produces a four-stage feature map, as the image encoder and a \unet{}-style decoder.  Also represented are both hierarchical and non-hierarchical features. Additionally, the components of the encoder-decoder network that are affected by pre-training and the components that are affected by the different downstream training strategies are labeled.}
    \label{fig:model_architecture}
\end{figure}

An \ac{fm} is a neural network pre-trained on a vast dataset that serves several distinct downstream tasks. 
The encoder-decoder architecture, illustrated in Figure \ref{fig:model_architecture}, serves as the basis of our framework developed to benchmark a broad variety of \acp{fm}.

We define an encoder-decoder model as the composition $\mathcal{D_{\theta_D} \circ E_{\theta_E}}$ of an encoder \eqref{eq:def_encoder} and a decoder \eqref{eq:def_decoder}.  Each of them is parametrized by parameters $\theta$ and together this composition approximates a real underlying function $f$ \eqref{eq:def_f}. This function maps from a two-dimensional space $X$ of height $H$ and width $W$ into another two-dimensional space $Y$ of the same height and width. While we assume two-dimensional input and output spaces of the same size, neither changing the sizes nor adding dimensions fundamentally alters the problem.

\begin{align}
    f&: X \in \mathbb{R}^{H \times W}  \mapsto Y \in \mathbb{R}^{H \times W} 
    \label{eq:def_f}
    \\
    \mathcal{E}_{\theta_E} &: X \in \mathbb{R}^{H \times W} \mapsto \{ E_s: E_s \in \mathbb{R}^{C_s \times H_s \times W_s} \quad \forall \quad 1 \leq s \leq S \}
    \label{eq:def_encoder}
    \\
     \mathcal{D}_{\theta_D} &:  \{ E_s: E_s \in \mathbb{R}^{C_s \times H_s \times W_s} \quad \forall \quad 1 \leq s \leq S \}  \mapsto Y^* \in \mathbb{R}^{H \times W} \text{, } Y^* \approx Y \in \mathbb{R}^{H \times W}
     \label{eq:def_decoder}
\end{align}

As defined in Equation \eqref{eq:def_encoder}, the encoder outputs a set of embeddings. Each embedding $E_s$ is an element of a three-dimensional space, with $C_s$ channels, a height of $H_s$ and a width of $W_s$. The total number of embeddings $S$ depends on the architecture and number of skip-connections. A neural network without skip-connections only uses one embedding and a neural network with skip-connections, such as a \unet, uses multiple embeddings. We define an encoder as non-hierarchical, if all embeddings have the same spatial dimensions, as defined in Equation \eqref{eq:def_hierarchical}. In contrast, a hierarchical model has embeddings at varying resolutions. 

\begin{align}
    \mathcal{E}_{\theta_E} &\coloneqq
    \begin{cases}
        \text{non-hierarchical}, & \text{if}\ H_i = H_j \quad \forall \quad 1 \leq i \leq j \leq S
        \\ 
        &  \land\ W_i = W_j \quad \forall \quad 1 \leq i \leq j \leq S
        \\
        \text{hierarchical}, & \text{otherwise}
    \end{cases} 
    \label{eq:def_hierarchical}
\end{align}

The choice of four embeddings ($S=4$), as illustrated in Figure \ref{fig:model_architecture}, is motivated by the fact that most hierarchical models, that we compared in this study, have exactly four distinct stages. 

Several pre-training methods exist. One possible pre-training method is self-supervised learning as outlined in Equation \eqref{eq:def_self_supervised}. 
We chose self-supervised learning as the example, because it is the most likely pre-training method for an \ac{sfm} and was already used by \cite{shengSeismicFoundationModel2024}, \cite{phamSeisBERTPretrainedSeismic2025} and \cite{sansalScalingSeismicFoundation2025}. 
In this context, self-supervised learning refers to \ac{mim} \cite[]{baoBEiTBERTPreTraining2022}, which is the most common technique within the self-supervised learning paradigm.

\begin{align}
    \text{self-supervised pre-training} &\coloneqq \min_{\theta^{PT}_E, \theta^{PT}_D} \sum_{i=0}^{N} L(X^*_i, X_i)
    \label{eq:def_self_supervised}
    \\
    &\coloneqq \min_{\theta_{E_{PT}}, \theta_{D_{PT}}} L_{ss}(\mathcal{D}_{\theta^{PT}_D}(\mathcal{E}_{\theta^{PT}_E}(X^M_i)), X_i)
    \nonumber
\end{align}

For self-supervised learning the mapping changes from $X \mapsto Y$ to $X^M \mapsto X$, where $X$ represents the original data, and $X^M$ denotes a randomly masked version of $X$.
Consequently,  $X^*$ is the prediction of the encoder-decoder network ($X^* \coloneqq \mathcal{D}_{\theta^{PT}_D}(\mathcal{E}_{\theta^{PT}_E}(X^M))$). 
Additionally, $L_{ss}$ is a pixel based loss, such as $\ell_{1}(X, Y)=||X-Y||_1$ or $\ell_{2}(X,Y)=||X-Y||_2$, with the specific loss depending on the implementation. 
It is important to note, that while an encoder-decoder network is used during pre-training, only the encoder is the \ac{fm}, as defined in Equation \eqref{eq:def_fm}. 
Therefore, the decoder is kept as small as possible ($|\theta^{PT}_D| \ll  |\theta^{PT}_E$|) shifting more responsibility to the encoder in order to create more meaningful features.

\begin{equation}
    FM_{\theta^{PT}_E}  \coloneqq Encoder_{\theta^{PT}_E} 
    \label{eq:def_fm}
\end{equation}

Pre-training with self-supervision can be done on real data. However, downstream training, as outlined in Equation \eqref{eq:def_supervised}, often relies on supervised learning, with $X$ representing the input, $Y$ the label, and $Y^*$ the prediction of the model ($Y^* \coloneqq \mathcal{D}_{\theta_D}(FM_{\theta_E | \theta^{PT}_E}(X))$). Similar to self-supervised learning, during supervised learning any pixel based loss can be used for $L_{s}$. However, based on some initial experiments and the findings by \cite{zhaoLossFunctionsImage2017}  we decided on the $\ell_{1}$ loss for our downstream training. 
\begin{align}
    \text{supervised downstream training} &\coloneqq \min_{\theta_E | \theta^{PT}_E, \theta_D} \sum_{i=0}^{N} L_{s}(Y^*_i, Y_i)
    \label{eq:def_supervised}
    \\
    &\coloneqq\min_{\theta_E | \theta^{PT}_E, \theta_D} \sum_{i=0}^{N} L_{s}(\mathcal{D}_{\theta_D}(FM_{\theta_E | \theta^{PT}_E}(X_i)), Y_i)
    \nonumber
\end{align}

Importantly, the downstream training is performed separately for each distinct task. For our downstream training, we use a decoder inspired by the right side of a \unet{} \cite[]{ronnebergerUNetConvolutionalNetworks2015}.
The \unet{} is a \ac{cnn} originally designed for biomedical image segmentation. It consists of an encoder-decoder structure with skip-connections. The encoder progressively reduces the spatial dimensions while increasing the feature depth, capturing context. The decoder then upsamples the feature maps to reconstruct the original image size, using skip connections from the encoder to retain high-resolution details. In our context, the encoder is replaced by an \ac{fm} and only the decoder part is kept, as illustrated in Figure \ref{fig:model_architecture} and defined in Equation \eqref{eq:def_supervised}. The four stage feature maps produced by the \ac{fm} are used as intermediate representations and then fed to the decoder. Therefore, the decoder has to adapt its channel size and upsampling factor to these feature maps.

The reason for choosing a \unet{} structure is its effectiveness in pixel-level regression tasks, such as the seismic processing tasks we consider. 
This robustness has been demonstrated multiple times, for example by \cite{fernandezDeepLearningSeismic2025} for demultiple, by \cite{fernandezComparisonDeepLearning2022a} and \cite{chaiDeepLearningIrregularly2020} for interpolation, by \cite{zhangUnsupervisedSeismicRandom2019} for denoising and by \cite{chengGenerativeFoundationModel2025} who used a \unet{} as basis for their \ac{gsfm}, among others. 

\subsection{Models}
\label{sec:models}

In order to select an architecture that is optimal for seismic processing tasks, a broad variety of \acp{fm} were compared. These \acp{fm} can be categorized into two groups: hierarchical and non-hierarchical models, as defined in Equation \eqref{eq:def_hierarchical}. In addition, we compared their architecture and their pre-training method. The architecture can either be convolutional-, transformer- or hybrid based  and the different pre-training methods are explained in the next section. All the models, compared in this study, as well as their characteristics are summarized in Table \ref{tab:models}. The selection of models was chosen to cover a broad range of distinct models with minimal overlap in their characteristics.

\begin{table}
    \centering
    \begin{tabular}{lllll}
        \toprule
        Model & Hierarchical & Architecture & Pre-Training & Citation \\ 
        \midrule
        \caformer{} \centeredcolorbox{CAFORMER}                         & yes   & Hybrid & SL \trainingshape{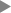}              & \cite{yuMetaFormerBaselinesVision2024a} \\
        \convnext{} \centeredcolorbox{CONVNEXT}                           & yes   & Convolutional & SL \trainingshape{supervised}              & \cite{liuConvNet2020s2022} \\
        \convnext{} \centeredcolorbox{CONVNEXT}                           & yes   & Convolutional & CL \trainingshape{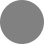}                     & \cite{liuConvNet2020s2022} \\
        \convnext{}(V2) \centeredcolorbox{CONVNEXT}                       & yes   & Convolutional & SSL \trainingshape{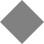}                     & \cite{wooConvNeXtV2Codesigning2023a} \\
        \acs{davit} \centeredcolorbox{DAVIT}                            & yes   & Transformer & MTL \trainingshape{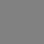}                     & \cite{dingDaViTDualAttention2022b} \\
        DeiT {\MakeUppercase{\romannumeral 3}} \centeredcolorbox{DEIT3}   & no    & Transformer & SL \trainingshape{supervised}              & \cite{touvronDeiTIIIRevenge2022} \\
        Dino \centeredcolorbox{DINO}                                  & no    & Transformer & SSLD \trainingshape{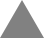}                   & \cite{oquabDINOv2LearningRobust2024a} \\
        \efficientnet{} \centeredcolorbox{EFFICIENTNET}                   & yes   & Convolutional & Semi-S \trainingshape{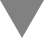}    & \cite{tanEfficientNetRethinkingModel2020} \\
        EVA02 \centeredcolorbox{EVA02}                                  & no    & Transformer & SSL \trainingshape{ss}                     & \cite{fangEVA02VisualRepresentation2024} \\
        EVA02 \centeredcolorbox{EVA02}                                  & no    & Transformer & CL \trainingshape{cl}                     & \cite{fangEVA02VisualRepresentation2024} \\
        \hiera{} \centeredcolorbox{HIERA}                             & yes   & Transformer & SSL \trainingshape{ss}                     & \cite{ryaliHieraHierarchicalVision2023} \\
        MambaOut \centeredcolorbox{MAMBAOUT}                               & yes   & Convolutional & SSL \trainingshape{ss}                     & \cite{yuMambaOutWeReally2024} \\
        \acs{swin} \centeredcolorbox{SWIN}                               & yes   & Transformer & SL \trainingshape{supervised}              & \cite{liuSwinTransformerHierarchical2021} \\
        \acs{swin} (V2) \centeredcolorbox{SWIN}                          & yes   & Transformer & SSL \trainingshape{ss}                     & \cite{liuSwinTransformerV22022} \\
        \vitamin{} \centeredcolorbox{VITAMIN}                           & yes   & Hybrid & CL \trainingshape{cl}                     & \cite{chenViTaminDesigningScalable2024} \\
        \bottomrule
    \end{tabular}
    \caption{Overview and categorization of the investigated models. For the pre-training strategy SL stands for supervised learning, SSL for self-supervised learning, SSLD for self-supervised learning with distillation, Semi-S for semi-supervised learning, MTL for multitask learning and CL for contrastive learning. The color next to the model denotes the model archetype, and in conjunction with the shape next to the pre-training strategy, this combination constitutes the legend for the majority of the plots in this study.}
    \label{tab:models}
\end{table}

\subsubsection{Pre-Training Methods}
\label{sec:pre_training}
The investigated models vary both in their architecture and in their pre-training method. Observed methods include supervised, self-supervised, self-supervised with distillation, semi-supervised, multitask and contrastive learning. Short introductions to each method are provided below, with further details available in their references.

Supervised learning is the most common training method and involves the use of labeled datasets and the prediction of these labels. In this context, this entails the utilization of a version of ImageNet \cite[]{russakovskyImageNetLargeScale2015} and the prediction of the class of each image. 

Self-supervised learning, in contrast to supervised-learning, utilizes unlabeled image datasets. In this context, a concrete form of self-supervised learning called \ac{mim} \cite[]{baoBEiTBERTPreTraining2022} is used. During training part of the image is masked, and the network has to recreate the image or predict the missing patches. The advantage is that unlabeled image datasets can be used. Furthermore, recreating the image is a more involved task than classification, which leads to more generalizable features \cite[]{heMaskedAutoencodersAre2021}. 

Self-supervised learning with distillation \cite[]{oquabDINOv2LearningRobust2024a} uses a teacher network, typically significantly larger than the student network, to teach the student to predict the outputs from the teacher. These outputs are a probability distribution over the target classes. By learning to predict the complete probability distribution of the teacher instead of just the correct label, the knowledge representation of the teacher model is distilled into the student model \cite[]{hintonDistillingKnowledgeNeural2015}. This is important because the smaller model may have insufficient capacity to learn this representation on their own. 

Semi-supervised training refers to the Noisy Student Training introduced in \cite{xieSelftrainingNoisyStudent2020}. It has three main steps: training a teacher on labeled images, using the teacher to generate pseudo labels on unlabeled data, training a student on a combination of labeled and unlabeled data. This process is iterated by switching the student to the teacher and training a new student.
This approach not only delivers better accuracy but also better generalization than supervised training \cite[]{xieSelftrainingNoisyStudent2020}. Additionally, similar to self-supervised training this method can benefit from vast unlabeled datasets. In contrast to self-supervised learning with distillation, only the pseudo labels are predicted and not the whole probability distribution of the teacher. A similar iterative approach, but without training multiple networks, was used by \cite{chengGenerativeFoundationModel2025} to shift their network from synthetic data to real data. 

Multitask learning references the training employed in \cite{xiaoFlorence2AdvancingUnified2024} and combines image-level, pixel-level and semantic tasks.  Image-level tasks involve comprehending the overall context of an image and the semantic relationships, with tasks like image classification, captioning, and visual question answering. Pixel level tasks focus on object and entity localization within images and understanding their relationships and spatial context, including object detection and segmentation. Semantic tasks require a detailed understanding of both text and image to align image regions with corresponding text phrases, challenging the model to grasp local details and semantic contexts. In contrast to the pre-training methods mentioned before, multitask learning is not image-to-image based, but requires a combination of image and text prompts as input and delivers a text prompt as output. Additionally, it requires an annotated dataset.

Finally, contrastive learning \cite[]{radfordLearningTransferableVisual2021} aims to learn a shared feature space for images and text. 
To achieve this a text encoder and image encoder are trained together to minimize the cosine similarity between correct (text, image) pairings and maximize the distance between incorrect pairings. 
However, similar to multitask learning, it requires an annotated dataset. 
On the other hand, this approach enables zero- and few-shot classification, not limited by the number of classes used during training. 

The majority of the models examined in this study have a corresponding pre-training method, with the exceptions of \convnext{} and \ac{swin}. Both are available pre-trained with supervision (V1) and self-supervision (V2), with only minimal architectural changes between them. Additionally, \convnext{} (in its V1 variant) is available with contrastive learning as pre-training strategy. Furthermore while DEIT3, DINO and EVA02 are distinct models, fundamentally they are all \acp{vit} with their the difference being their pre-training method. The corresponding pre-training method, as well as the architecture, for each tested model is summarized in Table \ref{tab:models}.

\subsection{Downstream Training Strategies}
\label{sec:training_strategies}

\begin{figure}
    \centering
    \includegraphics[width=\textwidth]{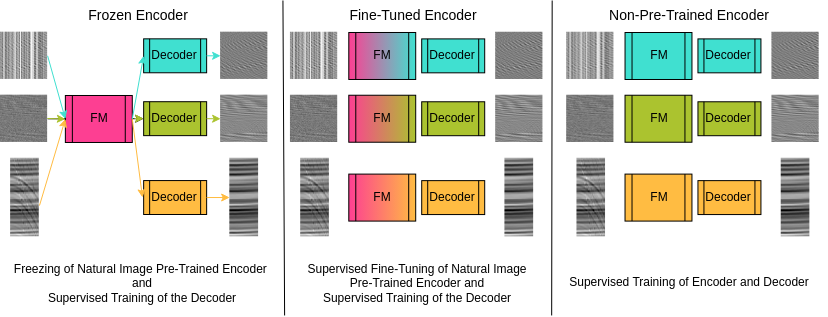}
    \caption{Comparison of the different downstream training strategies. \textit{Frozen encoder} freezes a natural image pre-trained \ac{fm} and only trains the decoder. \textit{Fine-tuned encoder} fine-tunes a natural image pre-trained \ac{fm} and trains the decoder. \textit{Non-pre-trained encoder} trains the whole encoder-decoder model from scratch. While all three downstream training strategies use supervised training, the pre-training method differs. Which specific pre-training method is used, is noted in Table \ref{tab:models}. Additionally, a short introduction about each observed pre-training method can be found in Section \ref{sec:training_strategies}.}
    \label{fig:training_strategies}
\end{figure}

In addition to the aforementioned models, this study compares three distinct downstream training strategies, as outlined in Figure \ref{fig:training_strategies}. These training strategies differ from the pre-training methods mentioned above. While the pre-training method defines how the \acp{fm}, which are used as image encoders, were pre-trained, the subsequent training strategies describe how the entire encoder-decoder model is trained for down-stream seismic applications. 

\begin{align}
    \text{frozen encoder} &\coloneqq \min_{\theta_D} \sum_{i=0}^{N} L_{s}(Y^*_i, Y_i)
    \label{eq:def_frozen_encoder}
    \\
    &\coloneqq  \min_{\theta_D} \sum_{i=0}^{N} L_{s}(\mathcal{D}_{\theta_D}(FM_{\theta^{PT}_E}(X_i)), Y_i)
    \nonumber
\end{align}

The first strategy, dubbed \textit{frozen encoder} employs \acp{fm} pre-trained on natural images directly (freezes the encoder) and only trains the decoder to adapt the output features of the \ac{fm} for seismic processing tasks. As described in Equation \eqref{eq:def_frozen_encoder}, freezing the encoder means that the parameters of the encoder stay untouched and are not further optimized. Analogue to Equation \eqref{eq:def_supervised},  $X$ is our input, $Y$ are our labels, $Y^*$ are the predictions of the encoder-decoder network and $L_{s}$ is the $\ell_1$ loss.

The second strategy, designated \textit{fine-tuned encoder}, uses the same \acp{fm} pre-trained on natural images. However, rather than freezing the pre-trained weights, it fine-tunes them, as well as training the decoder, as demonstrated in Equation \eqref{eq:def_pre_trained_encoder}.

\begin{align}
    \text{pre-trained encoder} &\coloneqq \min_{\theta_E | \theta^{PT}_E, \theta_D} \sum_{i=0}^{N} L_{s}(Y^*_i, Y_i)
    \label{eq:def_pre_trained_encoder}
    \\
    &\coloneqq  \min_{\theta_E | \theta^{PT}_E, \theta_D} \sum_{i=0}^{N} L_{s}(\mathcal{D}_{\theta_D}(FM_{\theta_E | \theta^{PT}_E}(X_i)), Y_i)
    \nonumber
\end{align}

The third strategy \textit{non-pre-trained encoder} also employs the same neural networks but without their pre-trained weights. Instead, the whole encoder-decoder is randomly initialized and then trained from scratch for each specific downstream task, as demonstrated in Equation \eqref{eq:def_non_pre_trained_encoder}.

\begin{align}
    \text{non-pre-trained encoder} &\coloneqq \min_{\theta_E, \theta_D} \sum_{i=0}^{N} L_{s}(Y^*_i, Y_i)
    \label{eq:def_non_pre_trained_encoder}
    \\
    &\coloneqq  \min_{\theta_E, \theta_D} \sum_{i=0}^{N} L_{s}(\mathcal{D}_{\theta_D}(\mathcal{E}_{\theta_E }(X_i)), Y_i)
    \nonumber
\end{align}

In most of the experiments, the strategy \textit{frozen encoder} was employed because it is significantly cheaper to compute. This is because the encoder, which corresponds to upwards of $80\%$ of the whole encoder-decoder model, is frozen and not trained. Our hypothesis is that the transfer learning capability of the network is indicative of the performance of the network for seismic pre-training. Furthermore, we aim to investigate the impact of natural image pre-training. Natural image datasets, unlike curated seismic data sets, are extensive and freely available. Furthermore, due to solely training the decoder the training takes hours instead of days, thereby enabling the assessment of a greater number of \acp{fm}. 

Given the substantial number of potential combinations of \acp{fm} and downstream tasks, it was not feasible to conduct individual hyperparameter optimizations for each combination. Instead, the experiments utilized a set of fixed hyperparameters, which included AdamW as the optimizer, a learning rate of $0.001$, a constant learning rate scheduler, and a weight decay of $0.01$.  The only difference of the hyperparameters between the downstream tasks was the number of epochs, with $50$ epochs selected for the demultiple task and $100$ epochs for the other two downstream tasks. This is due to latter two using a random cut augmentation and therefore diversifying the data more between epochs. Further details can be found in the source code.

%% file: sections/results.tex
\section{Results}
\label{sec:results}
In order to draw comparisons between the various \acp{fm}, it is necessary to consider a number of different dimensions. 
The primary and most significant of these is the architecture, which can be roughly divided into hierarchical and non-hierarchical models, as well as transformer-, convolution- and hybrid-based approaches. 
The classification of the models was presented earlier in Table \ref{tab:models}.
Another differentiating factor is the pre-training strategy, which is predominantly interesting from a transfer learning perspective and to a lesser extent to the training of a seismic processing \ac{fm}. 
This is due to the fact that the majority of pre-training strategies are not applicable to the seismic processing domain. 
A more relevant aspect, is the pre-training dataset size used, as this provides insight into the amount of data required to train an \ac{fm} for seismic processing. 
The final dimensions that will be examined in this study are model size and inference time. 
These are crucial because an \ac{sfm} must be efficient enough to handle the substantial amount of data involved in seismic processing.

\subsection{Quantitative Results}
\subsubsection{Experimental Setup and Data}
For the demultiple task a synthetic common depth point (CDP) sorted dataset is used. The seismic input gathers consist of the merged primaries and multiples, and the corresponding labels consist only of the primaries. Each gather has a size of $64 \times 512$ samples and the dataset consists of $100,000$ gathers. The synthetic seismic data was generated using the convolutional modeling as described in \cite{fernandezDeepLearningSeismic2025}. 

For the interpolation and denoising tasks, we use the open source 2007 BP Anisotropic Velocity Benchmark \cite[]{shah2007BPAnisotropic2007} dataset, consisting of 1641 shot gathers of size $800 \times 1151$. Due to the prominence of the first break and the extensive empty sections above it in the 2007 BP Anisotropic Velocity Benchmark, random cuts of the size $224 \times 224$ are applied below the first break. This is done, in order to ensure that each gather contains meaningful seismic information. For interpolation, we randomly mask the input gather and for denoising we add random uniform noise to the input gather. Details of further augmentation and pre-processing steps are available in the source code. 

In this study, each experiment recorded the \ac{psnr}, \ac{ssim} \cite[]{ssim_paper}, and \ac{mse} metrics. However, due to the high correlation among these metrics and a high correlation across the three downstream tasks only the combined \ac{ssim} (($SSIM_{combined}=SSIM_{Demultiple}+SSIM_{Interpolation}+SSIM_{Denoise}$)) is displayed. The \ac{ssim} was chosen for the combined score as it has a known upper limit of one, ensuring an equal weighting of the various downstream tasks and providing a theoretical upper bound for performance.

\subsubsection{Impact of the FM Architecture and Pre-Training Method}
\label{sec:base_models}

\begin{figure}
  \centering
  \includegraphics[width=\textwidth]{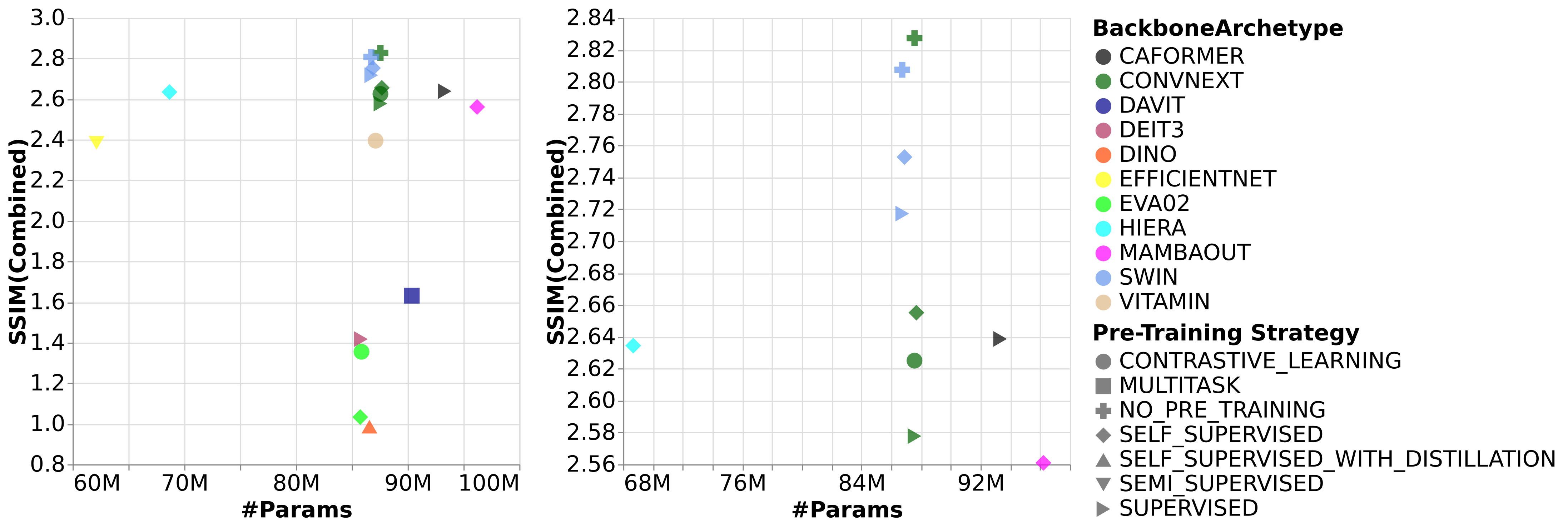}
  \caption{Comparison of the combined performance of the different \acp{fm} contrasted against their number of parameters. The left plot shows all the tested \acp{fm} and the right plot only the \acp{fm} with a combined \ac{ssim} above $2.5$ to further highlight the difference between the well performing models. The well performing models are \ac{swin}, \convnext, \hiera, \caformer{} and MambaOut. While most of the well performing models have a similar number of parameters \hiera{} is significantly smaller.}
  \label{fig:base_model_combined_parameters}
\end{figure}

Figure \ref{fig:base_model_combined_parameters} presents the combined \ac{ssim} for each model relative to the parameters of the \ac{fm} utilized as the image encoder. 
The legend includes the backbone archetype and pre-training methods. The archetype refers to a model family; for example, all \convnext{} models belong to the CONVNEXT archetype, regardless of pre-training strategy or version (V1 and V2). 
Besides the pre-training methods outlined previously, the legend also includes NO\_PRE\_TRAINING, indicating models trained with downstream training strategy \textit{non-pre-trained encoder}, see Equation\eqref{eq:def_non_pre_trained_encoder}.
These models serve as a baseline for comparison, while all other models were trained using the \textit{frozen encoder} strategy, see Equation \eqref{eq:def_frozen_encoder}. 
On the left, all tested models are shown; on the right, only models with a combined \ac{ssim} above 2.5 are displayed, excluding non-hierarchical models, as well as \efficientnet{} and \vitamin. 
The top-performing model is \ac{swin} (V2) pre-trained via self-supervision, followed by \ac{swin} (V1) pre-trained via supervision. 
The third-best is \convnext{} (V2) pre-trained via self-supervision, and the fourth-best is \caformer{}, closely followed by \hiera. 
Interestingly, \caformer{} performs well despite being trained via supervision. 
For both \ac{swin} and \convnext{}, the self-supervised variants surpass their supervised counterparts, suggesting that pre-training \caformer{} via self-supervision could significantly enhance results, potentially exceeding \convnext{} (V2).

\subsubsection{Impact of the Pre-Training Dataset}

\begin{figure}
  \centering
  \includegraphics[width=\textwidth]{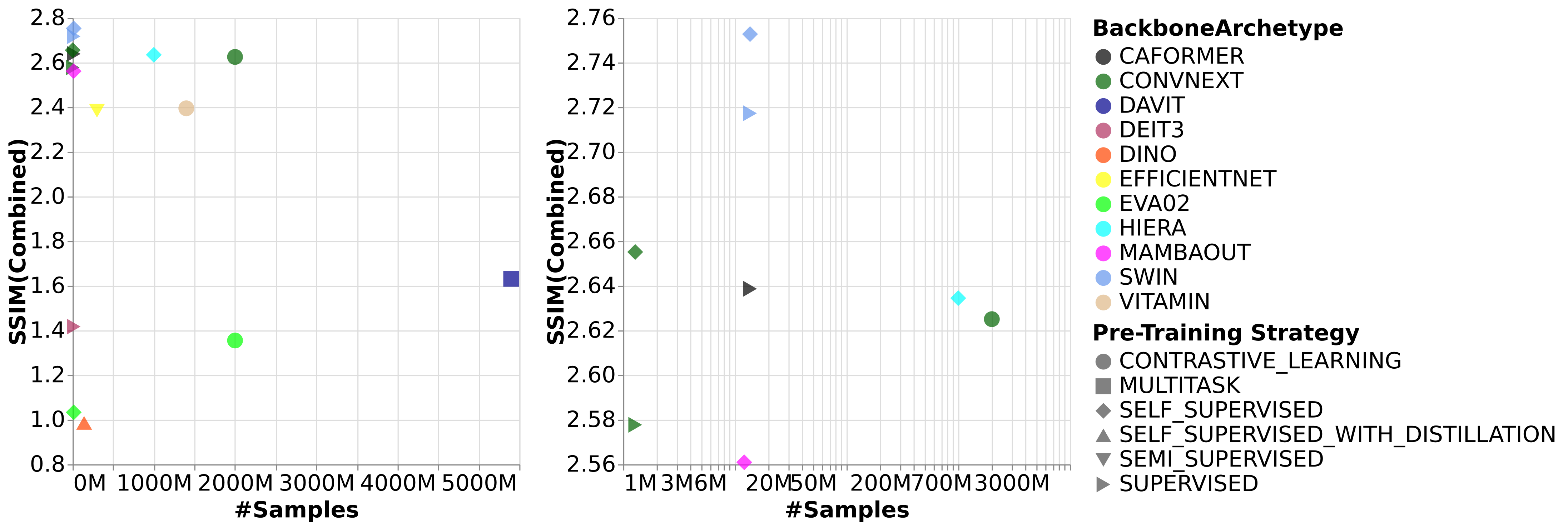}
  \caption{Comparison of the combined performance of the different \acp{fm} contrasted against the size of the pre-training dataset. The left plot shows all the tested \acp{fm} and the right one only the \acp{fm} with a combined \ac{ssim} above $2.5$ to further highlight the difference between the well performing models. Additionally, the right plot uses a log scaling for the x-axis to demonstrate the differences between the smaller datasets, which vanish with the linear scale. The well performing models are \ac{swin}, \convnext{}, \hiera{}, \caformer{} and MambaOut. Of these \hiera{} and MambaOut were trained on a significantly larger dataset and \convnext{} on the smallest.}
  \label{fig:base_model_combined_dataset_size}
\end{figure}

Figure \ref{fig:base_model_combined_dataset_size} reveals variations not only in pre-training strategies but also in the datasets and their sizes. 
Similar to Figure \ref{fig:base_model_combined_parameters}, all tested models are shown on the left, while only those with a combined \ac{ssim} above 2.5 are featured on the right. 
The right plot employs a logarithmic scale on the x-axis to better accentuate differences among smaller datasets, which might be obscured using a linear scale.

Although some non-hierarchical models were trained using extensive datasets, the architecture appears to be more crucial for seismic processing tasks. 
However, even among hierarchical models, significant differences in pre-training dataset sizes could influence the outcomes. 
While \ac{swin} surpasses \convnext{}, it also benefits from being pre-trained on a larger dataset (\imgnet{22} vs. \imgnet{1}). 
\hiera{} and \convnext{} trained via contrastive learning are exceptions, having been pre-trained on significantly larger datasets than the \ac{swin} models.

More interesting for an \ac{sfm} is, that models pre-trained on smaller datasets can still achieve impressive results.

\subsubsection{Impact of Model Size and Inference Time}

\begin{figure}
  \centering
  \includegraphics[width=\textwidth]{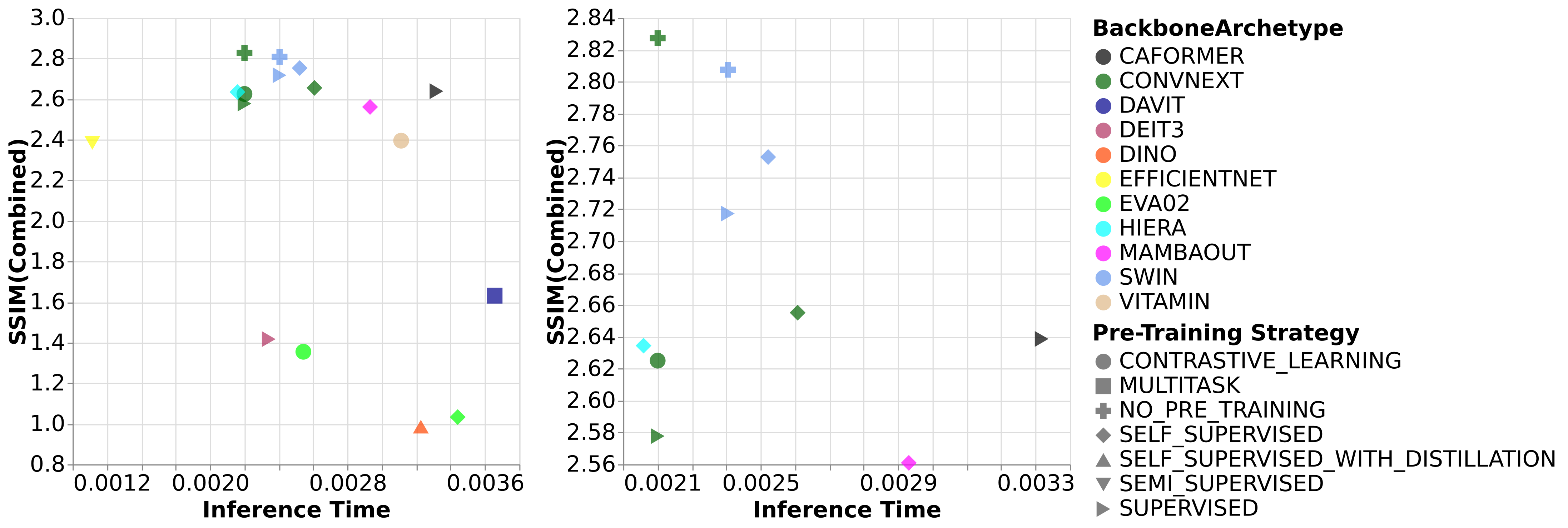}
  \caption{Comparison of the combined performance of the different \acp{fm} contrasted against their inference time. The left plot shows all the tested \acp{fm} and the right one only the \acp{fm} with a combined \ac{ssim} above $2.5$ to further highlight the difference between the well performing models. The well performing models are \ac{swin}, \convnext{}, \hiera{}, \caformer{} and MambaOut. Interestingly, the V2 versions of \ac{swin} and \convnext{} are slower than their V1 counterparts. Additionally, \caformer{} is significantly slower than the other testes \acp{fm}.}
  \label{fig:base_model_combined_inference_time}
\end{figure}

In this section, only the base model versions are compared, resulting in similar parameter counts, with two notable exceptions: \efficientnet{} and \hiera, which are substantially smaller, as depicted in Figure \ref{fig:base_model_combined_parameters}. 
While \efficientnet{} benefits from faster inference times, \hiera{} is only marginally faster than \convnext{}, despite having significantly fewer parameters, as illustrated in Figure \ref{fig:base_model_combined_inference_time}.
An interesting observation is that for both \convnext{} and \ac{swin}, the V2 versions—which incorporate minimal architectural changes for self-supervised pre-training—are slower than their V1 counterparts. 

Inference times were recorded on an A100 using the implementations from \cite{wightmanPyTorchImageModels2019}, without torch.compile, with a batch size of 512 and gathers of size 224x224.

The applicability of these larger \acp{fm} is demonstrated by the \ac{swin} (V2) inference time of approximately 2.5 milliseconds, translating to around 400 gathers per second. Notably, these inference results were obtained without any optimizations or quantization, which both could significantly enhance inference speed.

\subsubsection{Impact of natural image pre-training}

\begin{figure}
  \centering
  \includegraphics[width=0.5\textwidth]{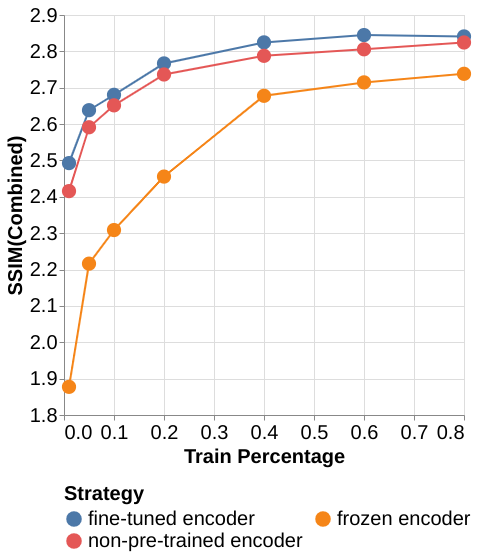}
  \caption{A comparison of the downstream training strategies for the \ac{swin} model family. The observed ranking of these downstream training strategies is \textit{fine-tuned encoder} $\geq$ \textit{non-pre-trained encoder} $\geq$ \textit{frozen encoder}, independently of the downstream task and dataset size.}
  \label{fig:comparison_training_strategies}
\end{figure}

Figure \ref{fig:comparison_training_strategies} provides a comparison of the downstream training strategies detailed previously for the \ac{swin} model family. 
The \ac{swin} models were selected due to their superior performance, as previously demonstrated. 
The observed hierarchy of downstream training strategies is \textit{fine-tuned encoder} $\geq$ \textit{non-pre-trained encoder} $\geq$ \textit{frozen encoder}, regardless of downstream task or dataset size. 
This indicates that pre-training on natural images is consistently comparable to or better than random initialization, especially in low data scenarios, though it holds true even for larger datasets. 
This suggests that pre-training on natural images may equip the model with capabilities that surpass what can be learned from the specific downstream task alone.

Moreover, the figure corroborates our hypothesis that the performance of the \textit{frozen encoder} strategy is linked to the other two strategies. 
Contrary to expectations, the performance gap with the \textit{frozen encoder} strategy diminishes as more task-specific training data becomes available. 
Initially, it was thought that pre-training would be beneficial in low-data contexts, as seen with the \textit{fine-tuned encoder} strategy. 
The need for more task-specific training data may be due to the domain gap between seismic data and natural images, suggesting that while features learned during pre-training are transferable, the decoder requires sufficient task-specific data to effectively leverage them.

\subsection{Qualitative Results}

This section provides qualitative results that complement the quantitative findings presented earlier. 
For brevity, the focus will be on the outcomes of two baseline models, \convnext{} (A) and \ac{swin} (B), alongside the top-performing non-baseline model, \ac{swin} V2 (C). 
Each model was specifically trained for its respective task, culminating in a total of nine unique configurations. 
The qualitative results are intended to contextualize the quantitative metrics rather than to assert superiority over the current state of the art.

\subsubsection{Interpolation}
\begin{figure}
  \centering
  \includegraphics[width=\textwidth]{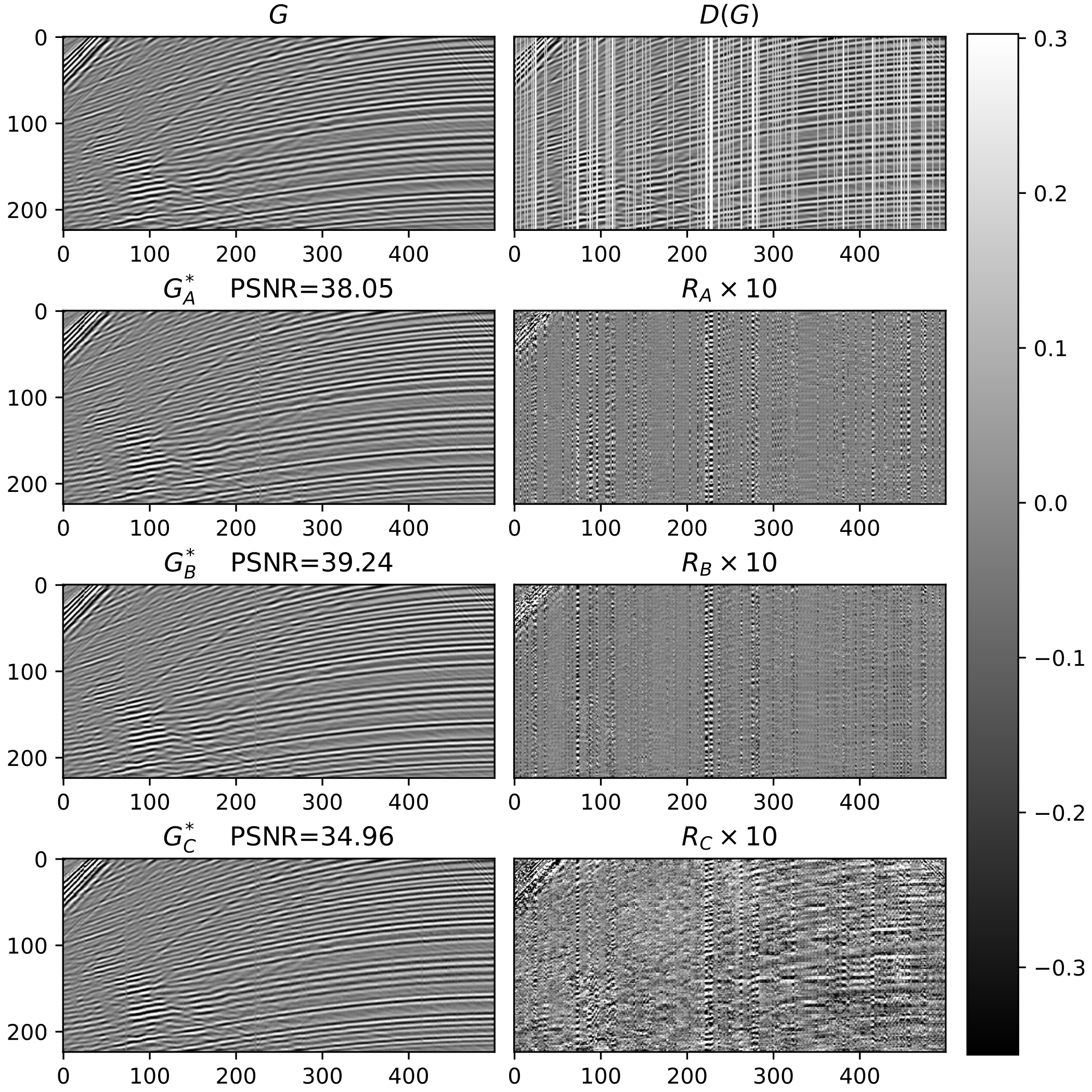}
  \caption{Interpolation results of the two baselines (\convnext{} (A) and \ac{swin} (B)) and the best overall model trained with strategy \textit{frozen encoder} (\ac{swin} V2 (C))  on a gather from the evaluation set. For each model the predicted gather $G^*$ as well as the residual $R=G-G^*$ are displayed.}
  \label{fig:interpolation_eval_set}
\end{figure}

Figure \ref{fig:interpolation_eval_set} depicts the results for the interpolation task  of each model. 
The input data comprises a gather from the evaluation set, which was not observed during training. 
The first row features this gather, denotes as $G$, alongside the decimated gather $D(G)$. Subsequent rows present the predicted gather $G^*$ and the residual, calculated as $R=(G-G^*) \times 10$, of each model. 
Although the predicted gathers ($P^*$) appear similar, the residuals ($R$) reveal more pronounced differences. 
The results of baseline (A) and baseline (B) show similarities, whereas  model (C) displays significant divergence, as confirmed by the \acp{psnr} values of $39.24$ for (B), $38.05$ for (A) and only $34.96$ for (C). 
Nonetheless, the results of (C) are impressive, considering that the encoder, which in this case accounted for $91\%$ of the total parameters, was trained exclusively on natural images. 

\subsubsection{Denoising}
\begin{figure}
  \centering
  \includegraphics[width=\textwidth]{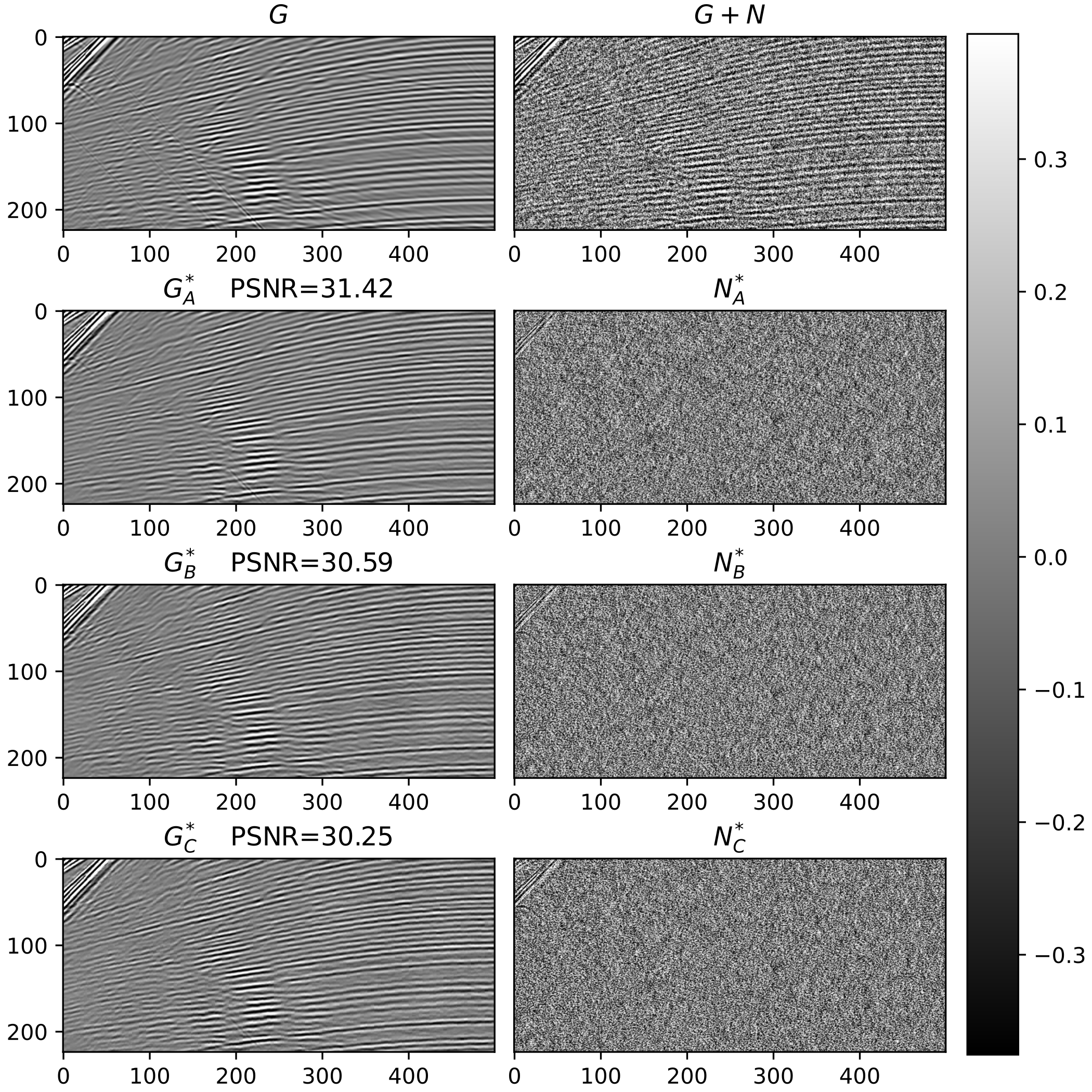}
  \caption{Denoising results of the two baselines (\convnext{} (A) and \ac{swin} (B)) and the best overall model trained with strategy \textit{frozen encoder} (\ac{swin} V2 (C))  on a gather from the evaluation set. For each model the predicted gather $G^*$ as well as the predicted noise $N^*=(G+N)-G^*$ are displayed.}
  \label{fig:denoise_eval_set}
\end{figure}

Figure \ref{fig:denoise_eval_set} presents the results of the three models for the denoising task. 
To evaluate the models' generalization capabilities, uniform noise is introduced to the input gather from the evaluation set, differing from the Gaussian noise used during training. 
The first row displays the gather $G$ and the noisy gather $G+N$. 
The subsequent rows consist of the predicted gather $G^*$ and the estimated noise $N^*=(G+N)-G^*$ for each model. 
In this instance, both the predicted gathers ($G^*$) and noise estimates ($N^*$) appear similar across all three models. 
This observation is supported by the \acp{psnr}, which are $30.59$ for (B), $31.42$ for (A), and $20.25$ for (C). 
Model (C)'s results are particularly impressive, as they exhibit quality comparable to the baselines, especially given that the encoder was not exposed to any seismic data.

\subsubsection{Demultiple}
\begin{figure}
  \centering
  \includegraphics[width=0.99\textwidth]{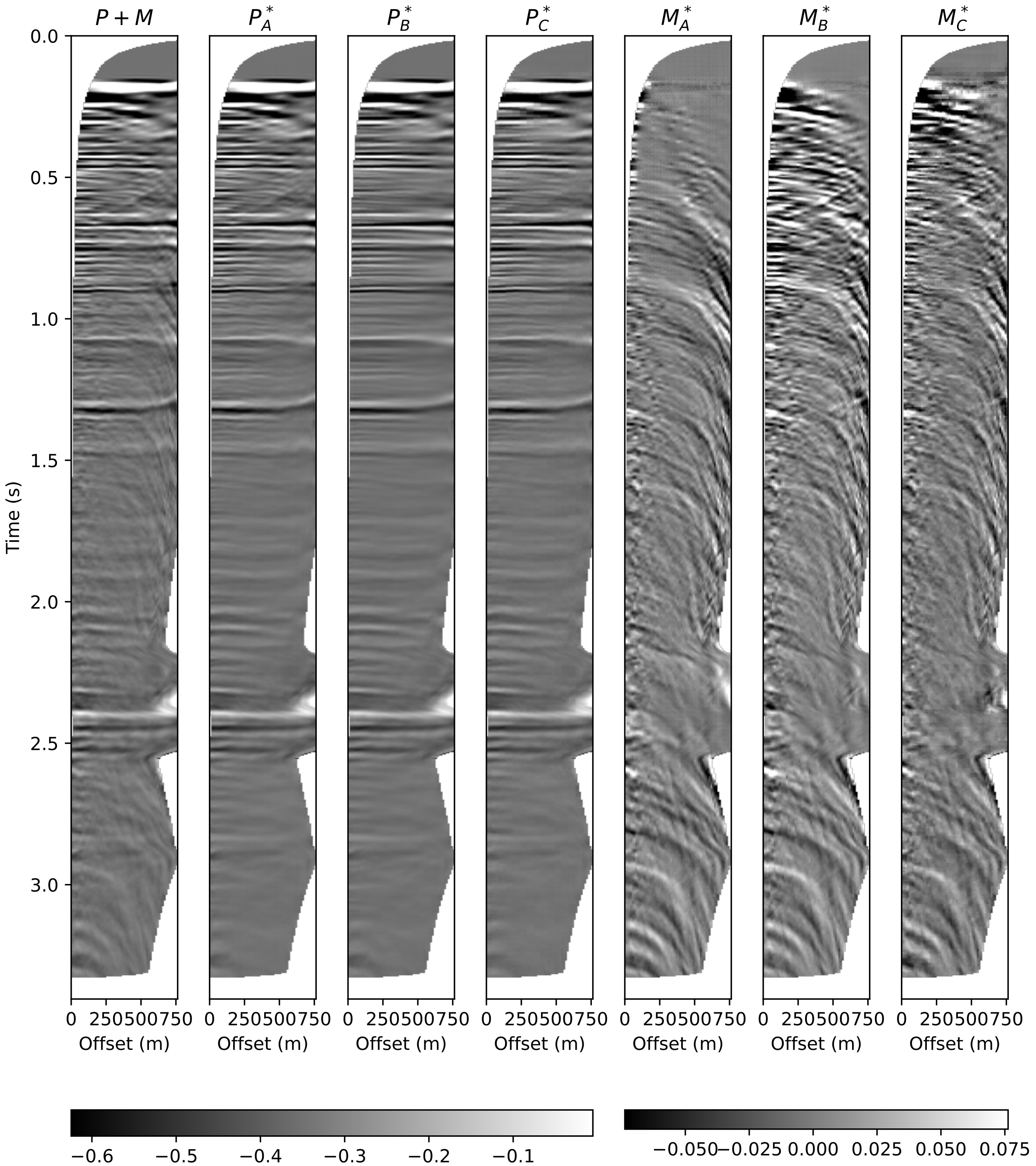}
  \caption{Demultiple results of the two baselines (\convnext{} (A) and \ac{swin} (B)) and the best overall model trained with strategy \textit{frozen encoder} (\ac{swin} V2 (C))  on  the North Sea Volve field data. For each model, the predicted primaries ($P^*$) and the predicted multiples ($M^*$) are displayed.}
  \label{fig:demultiple_field}
\end{figure}

Figure \ref{fig:demultiple_field} showcases the results for the demultiple task using a gather from a post-migration dataset from the North Sea Volve field. 
The first image shows the input gather containing primaries ($P$) and multiples ($M$). 
The following images depict the predicted primaries ($P^*$) from each model, while the last three images illustrate the predicted multiples ($M^*=(M+P)-P^*$), derived by subtracting the predicted primaries from the input gather.
While the predicted primaries ($P^*$) exhibit similarities, notable differences arise in the predicted multiples ($M^*$), particularly in the upper third, influenced by the mute and first break. 
Interestingly, the results of baseline (B) and model (C) are more alike than those of the two baseline models, suggesting that the architecture may play a more critical role than the training strategy in addressing the mute and first break. 
The performance of model (C) is especially remarkable, considering that only the decoder was trained on seismic data.

%% file: sections/discussion.tex
\section{Discussion}

In this section, we will explore several significant findings from our study. We will begin with the significance of hierarchical models, followed by the influence and importance of pre-training methods. Finally, we will address the issues of generalization and the performance of various downstream training strategies.

\subsection{Importance of hierarchical models}

The most critical finding in this paper is arguably the discrepancy in performance between hierarchical and non-hierarchical models. However, the majority of the \acp{fm} in the seismic space utilize a non-hierarchical model, as seen in \cite{shengSeismicFoundationModel2024}, \cite{sansalScalingSeismicFoundation2025} and \cite{phamSeisBERTPretrainedSeismic2025}, all of whom employ a \ac{vit}. It should be noted, that the results of this investigation are likely more pronounced due to the fact that all the downstream tasks are pixel-level regression tasks. The discrepancy between hierarchical and non-hierarchical models is likely diminished for downstream tasks that do not necessitate similar precision, such as salt body classification or fault detection.
Nonetheless, we believe that hierarchical models should still exhibit superior performance in these tasks, as corroborated by \cite{goldblumBattleBackbonesLargeScale2023}.

\subsection{Pre-training strategies and their relevance for an SFM}

The analysis of pre-training strategies is impeded by the lack of isolated comparison between them. Most pre-training strategies are only available for one model and one dataset. A more fair comparison would be interesting, but would need to be done in the natural image domain. This is because some of the pre-training strategies require a language mapping which is infeasible for seismic data. Additionally, most pre-training strategies require labeled data which is also infeasible for seismic data. Therefore, the only applicable pre-training strategy without synthetic data, for an \ac{sfm} is self-supervision. Relying on synthetic data semi-supervised learning could also be an alternative, as shown by \cite{chengGenerativeFoundationModel2025}.

A notable benefit of self-supervised pre-training on seismic data is its capacity to learn features directly from the data itself. This approach stands in contrast to the prevailing strategy of training on synthetic data and attempting to enhance it to improve generalization on real data. While Figure \ref{fig:comparison_training_strategies} demonstrates the comparable or superior performance of fine-tuning a natural image pre-trained model compared to only training on synthetic data, we hypothesize that pre-training on seismic data will yield even better results. A subject that merits further exploration is whether it is preferable to train smaller, more efficient models for significantly different regions or to train a single, large model on all available data.

\subsection{Baseline Performance and Generalization}

While the transfer performance of \convnext{} trained with strategy \textit{frozen encoder} was significantly worse than \ac{swin}, the baseline \convnext{} model that was trained using strategy \textit{non-pre-trained encoder} performed better than the \ac{swin} baseline. We hypothesize that this is due to these examples aligning closely to the training distribution and that \ac{swin} models generalize better. This is based on the quantitative results of these models trained with strategy \textit{frozen encoder} and further corroborated by the qualitative results for demultiple on the North Sea Volve field data.

On the topic of out-of-distribution generalization, we theorize that the training data used for interpolation and denoising while large enough for in-distribution generalization is too small for out-of-distribution generalization. 
Therefore, the qualitative results for these two task are shown for in-distribution examples. 
While the interpolation example is completely in-distribution, the denoise example is shifted a bit outside distribution by changing the noise distribution but not the underlying seismic distribution. 
In contrast, the qualitative example for demultiple is from the North Sea Volve field data and therefore out-of-distribution. 

%% file: sections/conclusion.tex
\section{Conclusion}
We carried out an extensive study on various \acp{fm} and developed a framework for further evaluation. 
Our primary focus was on seismic processing, assessing the performance of different \acp{fm} across three seismic processing tasks: demultiple, interpolation, and denoising. 
We quantitatively assessed a wide array of \acp{fm} using synthetic data to elucidate how different model attributes, such as pre-training strategies and architectural design choices, influence their performance. 
Furthermore, we selected a subset of models for qualitative analysis and presented demultiple results using real field data. 
Our comparison of different \acp{fm} also explored the impact of pre-training on natural images, with and without additional fine-tuning.

Identifying the optimal architecture for seismic processing is a complex and multifaceted challenge. 
However, our research has led to several key insights. 
First, future \acp{sfm}, particularly those intended for seismic processing tasks, should favor hierarchical models over non-hierarchical ones. 
Second, \ac{swin} appears to be the best overall model, followed by \convnext. 
Lastly, pre-training on natural images consistently performs as well as or better than training from scratch across all data scenarios.